\documentclass[runningheads]{llncs}
\usepackage[margin=1in]{geometry}
 
\usepackage{eccv}

\usepackage{multirow}
\usepackage{threeparttable}

\usepackage{eccvabbrv}

\usepackage{graphicx}
\usepackage{booktabs}

\usepackage[accsupp]{axessibility}  

\usepackage{hyperref}

\usepackage{orcidlink}

\begin{document}

\title{Aligning What EEG Can See: Structural Representations for Brain–Vision Matching} 


\author{
Jingyi Tang\inst{1}\thanks{Equal contribution} \and
Shuai Jiang\inst{1}\inst{*}\and
Fei Su\inst{1} \and
Zhicheng Zhao\inst{1}
}

\institute{Beijing University of Posts and Telecommunications, Beijing, 100876, China \\
\email{\{2024141108\}@bupt.cn}}

\maketitle
\setlength{\textwidth}{\dimexpr\paperwidth-3cm}


\begin{abstract}


    Visual decoding from electroencephalography (EEG) has emerged as a highly promising avenue for non-invasive brain-computer interfaces (BCIs). Existing EEG-based decoding methods predominantly align brain signals with the final-layer semantic embeddings of deep visual models. However, relying on these highly abstracted embeddings inevitably leads to severe cross-modal information mismatch. In this work, we introduce the concept of Neural Visibility and accordingly propose the EEG-Visible Layer Selection Strategy, aligning EEG signals with intermediate visual layers to minimize this mismatch. Furthermore, to accommodate the multi-stage nature of human visual processing, we propose a novel Hierarchically Complementary Fusion (HCF) framework that jointly integrates visual representations from different hierarchical levels. Extensive experiments demonstrate that our method achieves state-of-the-art performance, reaching an 84.6\% accuracy (+21.4\%) on zero-shot visual decoding on the THINGS-EEG dataset. Moreover,  our method achieves up to a 129.8\% performance gain across diverse EEG baselines, demonstrating its robust generalizability.

\end{abstract}

\section{Introduction}
\label{sec:intro}
Understanding how human visual representations are reflected in neural activity is a central objective in cognitive neuroscience and brain–computer interface (BCI) research. In recent years, with the rapid development of deep visual models, researchers have increasingly adopted outputs from pretrained visual networks as intermediate representations. EEG signals are mapped into these visual feature spaces to enable EEG-to-image retrieval, neural decoding, and brain-driven image generation. In particular, contrastive learning–based cross-modal alignment frameworks, often built upon large-scale vision–language models such as CLIP, have become the dominant approach in this field\cite{li2024visual,Wu2025BridgingTV,song2023decoding,du2023decoding}.

However, we argue that existing paradigms overlook a fundamental constraint: {\bf the neural visibility of EEG signals}.

In this work, we define the neural visibility of EEG signals as the property that, during perception, visual information can not only be 
encoded into EEG signals but also be reliably decoded by data-
driven models.

Neurophysiological studies and EEG-based neural decoding research indicate that the neural visibility of different visual components in EEG is not uniform\cite{bar2006top,petras2019coarse,holm2024contribution,man1982computational}. Specifically, EEG responses to {\bf high spatial frequency (HSF)} information, corresponding to fine-grained visual details, relatively weak and more susceptible to noise interference\cite{petras2019coarse,man1982computational,goffaux2011coarse}. In contrast, {\bf low spatial frequency (LSF) } information, which conveys global structural content, elicits more stable and robust responses \cite{bar2006top,holm2024contribution,goffaux2011coarse}. Moreover, {\bf high-level semantic information} is typically represented in EEG in a more indirect manner, relying on later-stage cognitive processing and being strongly affected by task demands and inter-subject variability, resulting in comparatively lower neural visibility\cite{paz2006electrophysiological,qi2017native}.

Deep visual models naturally exhibit a hierarchical organization of information. 
Early layers primarily capture local edges and textures associated with high spatial frequency (HSF) details. 
Intermediate layers progressively integrate larger receptive fields, forming representations of object shape, contours, and part relationships that align more closely with low spatial frequency (LSF) structural information. 
Deeper layers further abstract these features into high-level semantic representations\cite{bau2017network,zeiler2014visualizing}. 
Existing research typically treats the final-layer embeddings of deep visual models, such as CLIP, as the optimal alignment target. 
This choice is mainly driven by their strong semantic expressiveness and cross-view invariance.

However, the final-layer representations of vision-language models are trained through large-scale image-text alignment to maximize semantic consistency. As a result, they often suppress fine-grained texture details and specific structural variations \cite{radford2021learning,jing2024fineclip}. Such an alignment strategy overlooks an important constraint: EEG signals exhibit low neural visibility for high-level semantic information represented in the final layers of visual models, while showing higher and more stable visibility for global structural information integrated in intermediate layers, which is predominantly driven by low spatial frequencies. Ignoring this constraint not only limits the effectiveness of cross-modal alignment, but also prevents models from adaptively weighting visual components according to their differential neural visibility in EEG.

Based on these observations, we argue that cross-modal alignment should prioritize visual information with higher neural visibility in EEG. Therefore, instead of aligning EEG with conventional final-layer semantic embeddings, we propose an {\bf EEG-Visible Layer Selection Strategy}, which performs cross-modal alignment with intermediate-layer embeddings of deep visual models. This strategy reduces cross-modal information mismatch and improves the stability and robustness of neural decoding. Furthermore, considering that EEG signals may integrate information from multiple stages of visual processing, we propose a {\bf Hierarchically Complementary Fusion (HCF) }framework. This framework jointly combines visual representations from different layers to better align with the information that is neurally visible in 
EEG.



Experimental results on multiple EEG-to-image retrieval tasks demonstrate that both the EEG-Visible Layer Selection Strategy and the Hierarchically Complementary Fusion framework consistently outperform conventional approaches that rely solely on final-layer representations. The results of frequency decomposition experiments indicate that
structure-dominated information driven by low spatial frequencies exhibits higher stability in EEG and stronger cross-condition generalization potential.


In summary, our contributions are as follows:



\begin{itemize}
    \item
    We introduce the concept of Neural Visibility and accordingly propose the EEG-Visible Layer Selection Strategy, aligning EEG signals with intermediate visual layers to minimize cross-modal mismatch.
    \item 
    We propose a novel Hierarchically Complementary Fusion (HCF) framework, which jointly combines visual representations from different layers to better align with the multi-stage nature of human visual processing.
    \item 
    Extensive quantitative and qualitative experiments demonstrate SOTA performance, achieving an 84.6\% accuracy (+21.4\%) on zero-shot visual decoding and up to a 129.8\% gain across baseline EEG encoders, proving its strong generalizability.
\end{itemize}

\section{Related Work}
\subsection{Spatial Frequency Information in Images}
Natural images contain information distributed across different spatial frequencies, which is commonly divided into low spatial frequency and high spatial frequency components. A large body of neuroscience research\cite{song2023decoding,bar2006top,man1982computational,bullier2001integrated} shows that human visual processing follows a hierarchical coarse-to-fine pathway. Specifically, the brain tends to extract global LSF structure first and then resolves fine-grained HSF details. Prior studies\cite{man1982computational,merigan1993parallel} further suggest that LSF information has a shorter processing latency, helping form an overall representation of a visual scene at early stages of perception. Petras \etal. \cite{petras2019coarse} reported a guiding role of LSF: it not only facilitates subsequent processing of HSF information, but may also suppress HSF-related processing. In addition, multiple studies\cite{man1982computational,goffaux2011coarse} found that, compared with HSF, LSF representations are more stable and more robust to noise, making them a highly reliable source of information for neural decoding. Motivated by these findings, our framework prioritizes aligning low-frequency neural features to better cope with the intrinsic noise in EEG signals.

\subsection{Multimodal Contrastive Representation Learning}
With the emergence of contrastive learning paradigms such as CLIP \cite{radford2021learning}, ALIGN \cite{jia2021scaling}, and BLIP \cite{li2022blip}, cross-modal pretraining has become a fundamental foundation for multimodal learning. These approaches have achieved remarkable progress in tasks such as image-text retrieval, zero-shot recognition, and image classification.
However, in EEG-image retrieval, cross-modal alignment becomes substantially challenging due to the low signal-to-noise ratio of EEG signals and the significant representational gap between neural activity and visual features. These factors limit both retrieval performance and cross-subject generalization.

\subsection{Neural Visual Decoding}
Recently, several studies have attempted to establish more stable cross-modal correspondences between EEG and image modalities. Wu \etal \cite{Wu2025BridgingTV} proposed an Uncertainty-aware Blur Prior (UBP), which dynamically suppresses fine image details to mitigate modality mismatch caused by both systematic and random differences between EEG signals and visual stimuli. In addition, Zhang \etal. \cite{Zhang2025NeuroBridgeBS} introduced the NeuroBridge framework, which combines Cognitive Prior Augmentation (CPA)—including Gaussian blur, noise, low-resolution degradation, and mosaic transformations—with a shared semantic projector, significantly improving zero-shot retrieval performance.
The image augmentation strategies used in these works have been shown to effectively improve retrieval performance. Although they are often interpreted as simulating perceptual uncertainty or reducing cross-modal mismatch, these operations also systematically alter the frequency distribution of images. In particular, they tend to attenuate high-frequency information while relatively preserving low-frequency structures. This observation suggests that, at the physical signal level, EEG responses may have higher visibility to low-frequency visual information.

\subsection{Multi-layer Fusion}
Multi-layer feature aggregation has been widely explored in cross-modal vision–language research. For example, FPN \cite{lin2017feature} fuses high-level semantic information with low-level localization cues across multiple scales to build a stable feature pyramid, which helps reduce the semantic gap between modalities. In addition, HyperNet \cite{kong2016hypernet} shows that combining features from layers with larger hierarchical gaps often performs better than combining adjacent layers, suggesting that appropriate layer selection is more important than simply aggregating more layers.

Since EEG signals naturally have a low signal-to-noise ratio and contain information across multiple temporal and spatial scales, similar ideas have also been explored in EEG decoding tasks. For example, Amin et al. \cite{amin2019multilevel} extract features from multiple convolutional layers of a CNN, assign weights to each layer, and fuse them for final prediction. This approach achieves better performance than using only the final-layer features.
Other studies \cite{amin2020multi,yang2024novel} further combine multi-layer CNN features using autoencoders or attention mechanisms to obtain representations that are more suitable for EEG data, thereby improving decoding accuracy.
These studies suggest that EEG representations are not limited to a single level of information. Instead, multi-layer modeling and fusion can better capture the complex structure embedded in brain signals.

\section{Method}
\subsection{Problem Formulation}
This study focuses on cross-modal alignment between EEG signals and visual feature representations. Given a dataset of paired EEG recordings and their corresponding images:
\begin{equation}
D = (I^{(n)}, E^{(n)} )_{n=1}^{N}
\end{equation}
the multi-channel EEG signal is denoted as $E \in \mathit{R}^{C_E \times T}$, where $\mathit{C_E}$ represents the number of EEG channels and $T$ denotes the temporal length of the signal. The corresponding visual stimulus image is denoted as $E \in \mathit{R}^{C_I \times H \times W}$, where $\mathit{C_I}$ is the number of image channels, with spatial dimensions $H \times W$. 

The objective of cross-modal alignment is to learn a shared embedding space in which matched EEG-image pairs are mapped closer together, while mismatched pairs are pushed farther apart. This enables EEG-to-image retrieval and neural decoding tasks.

\subsection{Cross-modal Alignment Framework}
\begin{figure}[t]
  \centering
  \includegraphics[width=0.8\textwidth]{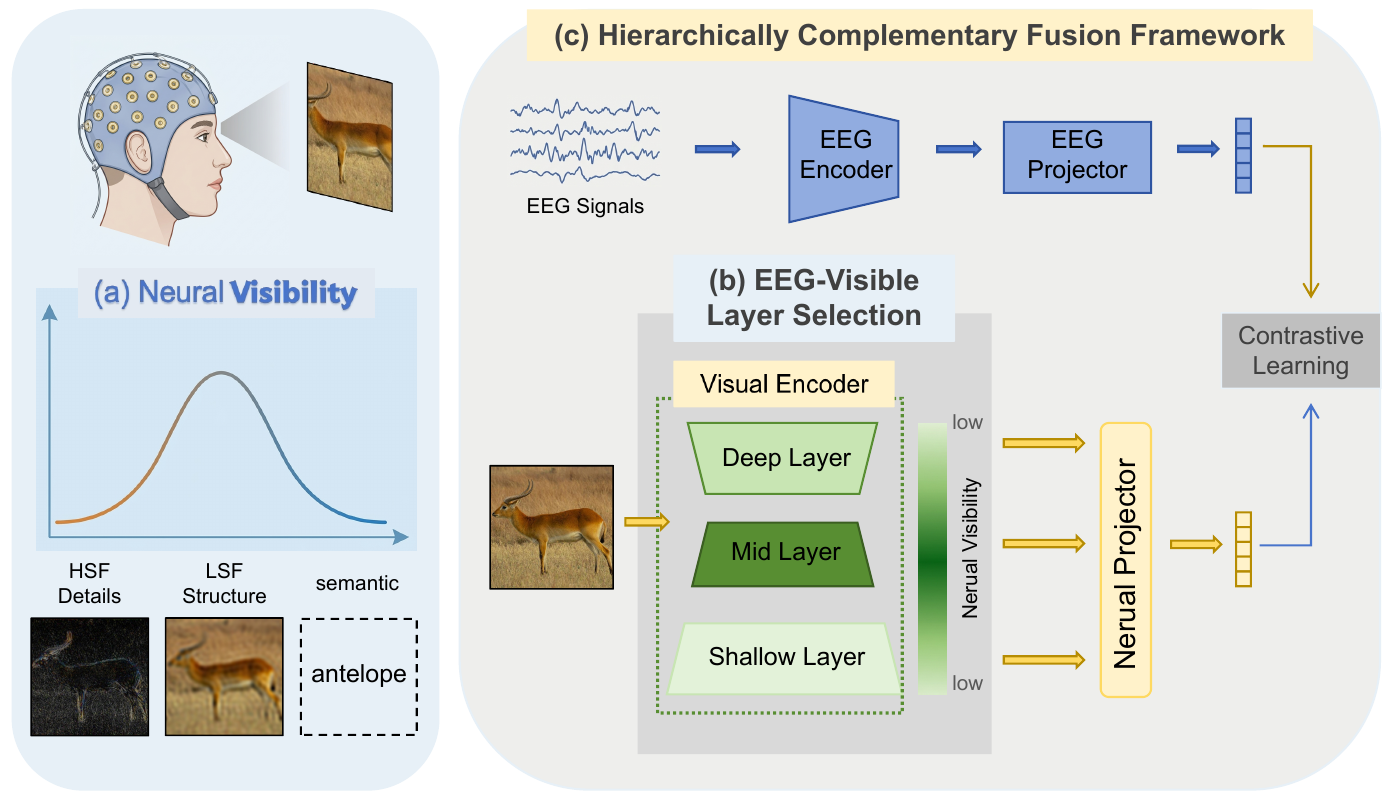}
  \caption{Overview of the proposed EEG-visible alignment framework for brain–vision matching.}
  \label{fig:main}
\end{figure}

We construct a cross-modal joint alignment framework to learn consistent representations between EEG signals and visual images, as illustrated in \cref{fig:main}. 

First, inspired by the asymmetric augmentation strategy proposed in NeuroBridge\cite{Zhang2025NeuroBridgeBS}, we apply a single augmentation $t_{E}$ to EEG data to mitigate noise interference:
\begin{equation}
X_E = t_E(E)
\end{equation}
At the same time, we apply $K$ augmentation transformations $t_{I,k}$ to the image data to attenuate high-frequency details, encouraging the model to rely more on global structural information associated with low spatial frequencies:
\begin{equation}
X_{I,k} = t_{I,k}(I), \quad k \in \{1, 2, \ldots, K\}
\end{equation}

Subsequently, the augmented EEG data $X_{E}$ is fed into an EEG encoder to extract EEG feature representations. The augmented image data $X_{I}=\{X_{I,k}\}$ is processed by a frozen pre-trained CLIP visual encoder to obtain image features.

Next, trainable linear projection heads are applied to map the EEG features and image features into a shared semantic embedding space, enabling cross-modal alignment:
\begin{equation}
z_E = P_E\!\left(f_E(X_E)\right), \quad
z_I = P_I\!\left(f_I(X_I)\right)
\end{equation}
where $f_E(\cdot)$ and $f_I(\cdot)$ denote the EEG encoder and the frozen CLIP visual encoder, and $P_E(\cdot)$ and $P_I(\cdot)$ represent linear projection functions.

Finally, in the shared embedding space, a contrastive objective is employed to reduce the distributional gap between EEG and image features, thereby enabling effective cross-modal alignment.
\begin{equation}
\begin{aligned}
\mathcal{L}_{\mathrm{InfoNCE}}
&=
-\frac{1}{2N}\sum_{i=1}^{N}
\Bigg[
\log
\frac{\exp\!\left(\mathrm{sim}(\mathbf{z}_E^{(i)},\mathbf{z}_I^{(i)})/\tau\right)}
{\sum_{j=1}^{N}\exp\!\left(\mathrm{sim}(\mathbf{z}_E^{(i)},\mathbf{z}_I^{(j)})/\tau\right)} \\
&\quad \log
\frac{\exp\!\left(\mathrm{sim}(\mathbf{z}_I^{(i)},\mathbf{z}_E^{(i)})/\tau\right)}
{\sum_{j=1}^{N}\exp\!\left(\mathrm{sim}(\mathbf{z}_I^{(i)},\mathbf{z}_E^{(j)})/\tau\right)}
\Bigg].
\end{aligned}
\end{equation}

\subsection{EEG-Visible Layer Selection Strategy}
\label{sec:layer_selection}
According to the proposed EEG-Visible Alignment Hypothesis, non-invasive EEG does not capture visual information in a complete or uniform manner. Based on this hypothesis, we select intermediate-layer outputs from the visual encoder—whose representations exhibit stronger consistency with EEG features—as the image features for alignment. This design facilitates more effective alignment within the shared feature space.

During forward propagation, the visual encoder $f_I(\cdot)$ produces a sequence of layer-wise feature activations:
\begin{equation}
F^{(l)} = f_{\mathrm{I}}^{(l)}(X_{I}), \quad l = 1, \ldots, L
\end{equation}
To convert the layer-wise feature activations into alignment-ready feature vectors, we design different pooling strategies according to the architectural characteristics of each backbone.
\begin{equation}
v^{(l)} = \operatorname{Pool}(F^{(l)})
\end{equation}
\subsubsection{ResNet Architecture.}
The feature activations at the $l$-th layer are denoted as $F^{(l)} \in \mathbb{R}^{C_l \times H_l \times W_l}$, where $C_l$ is the number of channels, and $H_l$ and $W_l$ are the spatial height and width of the feature map. 
We apply global average pooling and global maximum pooling separately to each layer to extract global representations with different statistical characteristics.

\subsubsection{Vision Transformer Architecture.} The feature activations at the $l$-th layer are represented as $F^{(l)} \in \mathbb{R}^{(N+1) \times D}$, where $N$ is the number of image patches, the additional $1$ token corresponds to the CLS token representing global information, and $D$ is the token embedding dimension. 
Similarly, we adopt two pooling strategies: CLS token pooling and mean pooling.

\subsection{Hierarchically Complementary Fusion (HCF) Framework}
Prior studies\cite{cichy2016comparison} have shown that human visual processing unfolds in temporally staged and spatially hierarchical phases. As a result, neural signals inherently contain information from multiple stages of visual processing. Motivated by this observation, we propose a Hierarchically Complementary Fusion (HCF) Framework. This framework employs a block-wise projection mechanism that enables the model to adaptively adjust the contributions of different visual layers according to the physiological visibility of EEG signals.

First, the pooled feature vectors from different layers, $\mathbf{v}_i \in \mathbb{R}^{d_i}$, are concatenated to form a fused visual representation:
\begin{equation}
\mathbf{v}_{\text{fuse}} =
\begin{bmatrix}
\mathbf{v}_1 \\
\mathbf{v}_2 \\
\vdots \\
\mathbf{v}_k
\end{bmatrix}
\in \mathbb{R}^{\sum_{i=1}^{k} d_i}.
\end{equation}

Next, a linear projection function $P(\cdot)$ maps the fused representation into the shared embedding space:
\begin{equation}
\hat{\mathbf{v}} = g_F(\mathbf{v}_{\text{fuse}}) 
= W_F \mathbf{v}_{\text{fuse}} + \mathbf{b}.
\end{equation}
The learnable weight matrix can be decomposed as
\begin{equation}
W_F = [W_1 \; W_2 \; \cdots \; W_k],
\end{equation}
where $W_i \in \mathbb{R}^{d \times d_i}$ corresponds to the weight assigned to the $i$-th layer feature. 
The projection can therefore be rewritten as:
\begin{equation}
\hat{\mathbf{v}} = \sum_{i=1}^{k} W_i \mathbf{v}_i + \mathbf{b}.
\end{equation}

This adaptive weighting mechanism enables dynamic adjustment of layer contributions. 
During training, the contrastive loss (\eg InfoNCE loss) implicitly optimizes the contribution of each visual layer, allowing the model to integrate complementary information across hierarchical levels and emphasize components that are more consistent with EEG representations.

\begin{table*}[!t]
\centering
\caption{Top-1/Top-5 accuracy (\%) for 200-way zero-shot retrieval on THINGS-EEG}
\label{tab:main_results}

\footnotesize   
\setlength{\tabcolsep}{4pt}  

\begin{tabular}{l l cccccccccc c}
\toprule
Method &  & Sub1 & Sub2 & Sub3 & Sub4 & Sub5 & Sub6 & Sub7 & Sub8 & Sub9 & Sub10 & Avg \\
\midrule
\multicolumn{13}{c}{\bf{Intra-Subject: train and test on one subject}} \\
\midrule

BraVL & Top-1 & 6.1 & 4.9 & 5.6 & 5.0 & 4.0 & 6.0 & 6.5 & 8.8 & 4.3 & 7.0 & 5.8 \\
      & Top-5 & 17.9 & 14.9 & 17.4 & 15.1 & 13.4 & 18.2 & 20.4 & 23.7 & 14.0 & 19.7 & 17.5 \\
\midrule

NICE  & Top-1 & 13.2 & 13.5 & 14.5 & 20.6 & 10.1 & 16.5 & 17.0 & 22.9 & 15.4 & 17.4 & 16.1 \\
      & Top-5 & 39.5 & 40.3 & 42.7 & 52.7 & 31.5 & 44.0 & 42.1 & 56.1 & 41.6 & 45.8 & 43.6 \\
\midrule

ATM   & Top-1 & 25.6 & 22.0 & 25.0 & 31.4 & 12.9 & 21.3 & 30.5 & 38.8 & 34.4 & 29.1 & 27.1 \\
      & Top-5 & 60.4 & 54.5 & 62.4 & 60.9 & 43.0 & 51.1 & 61.5 & 72.0 & 51.5 & 63.5 & 58.1 \\
\midrule

CognitionCapturer & Top-1 & 27.2 & 28.7 & 37.2 & 37.7 & 21.8 & 31.6 & 32.8 & 47.6 & 33.4 & 35.1 & 33.3 \\
                  & Top-5 & 59.5 & 57.0 & 66.1 & 63.2 & 47.8 & 58.1 & 59.6 & 73.5 & 57.7 & 63.6 & 60.6 \\
\midrule               
Neural-MCRL & Top-1 & 27.5 & 28.5 & 37.0 & 35.0 & 22.5 & 31.5 & 31.5 & 42.0 & 30.5 & 37.5 & 32.4 \\
             & Top-5 & 64.0 & 61.5 & 69.0 & 66.0 & 51.5 & 61.0 & 62.5 & 74.5 & 59.5 & 71.0 & 64.1 \\   
\midrule
VE-SDN & Top-1 & 32.6 & 34.4 & 38.7 & 39.8 & 29.4 & 34.5 & 34.5 & 49.3 & 39.0 & 39.8 & 37.2 \\
        & Top-5 & 63.7 & 69.9 & 73.5 & 72.0 & 58.6 & 68.8 & 68.3 & 79.8 & 69.6 & 75.3 & 70.0 \\
\midrule
UBP   & Top-1 & 41.2 & 51.2 & 51.2 & 51.1 & 42.2 & 57.5 & 49.0 & 58.6 & 45.1 & 61.5 & 50.9 \\
      & Top-5 & 70.5 & 80.9 & 82.0 & 76.9 & 72.8 & 83.5 & 79.9 & 85.8 & 76.2 & 88.2 & 79.7 \\
\midrule

NeuroBridge
      & Top-1 & 50.0 & 63.2 & 61.6 & 61.4 & 54.8 & 69.7 & 62.7 & 71.2 & 64.0 & 73.6 & 63.2 \\
      & Top-5 & 77.6 & 90.6 & 91.1 & 90.0 & 85.0 & 92.9 & 88.8 & 95.1 & 91.0 & 97.1 & 89.9 \\
\midrule

\bf{HCF(Ours)}
      & Top-1 & \bf 81.9 & \bf 86.0 & \bf 84.3 & \bf 82.2 & \bf 74.5 & \bf 90.4 & \bf 82.8 & \bf 90.9 & \bf 79.7 & \bf 93.1 & \bf 84.6 \\
      & Top-5 & \bf 98.3 & \bf 98.9 & \bf 97.7 & \bf 98.8 & \bf 94.1 & \bf 98.9 & \bf 97.2 & \bf 99.4 & \bf 98.4 & \bf 99.8 & \bf 98.2 \\
\midrule
\multicolumn{13}{c}{\bf{Inter-Subject: leave one subject out for test}} \\
\midrule

BraVL & Top-1 & 2.3 & 1.5 & 1.4 & 1.7 & 1.5 & 1.8 & 2.1 & 2.2 & 1.6 & 2.3 & 1.8 \\
      & Top-5 & 8.0 & 6.3 & 5.9 & 6.7 & 5.6 & 7.2 & 8.1 & 7.6 & 6.4 & 8.5 & 7.0 \\
\midrule

NICE  & Top-1 & 7.6 & 5.9 & 6.0 & 6.3 & 4.4 & 5.6 & 5.6 & 6.3 & 5.7 & 8.4 & 6.2 \\
      & Top-5 & 22.8 & 20.5 & 22.3 & 20.7 & 18.3 & 22.2 & 19.7 & 22.0 & 17.6 & 28.3 & 21.4 \\
\midrule

ATM      & Top-1 & 10.5 & 7.1 & 11.9 & 14.7 & 7.0 & 11.1 & 16.1 & 15.0 & 4.9 & 20.5 & 11.9 \\
         & Top-5 & 26.8 & 24.8 & 33.8 & 39.4 & 23.9 & 35.8 & 43.5 & 40.3 & 22.7 & 46.5 & 33.8 \\
\midrule

UBP      & Top-1 & 11.5 & 15.5 & 9.8 & 13.0 & 8.8 & 11.7 & 10.2 & 12.2 & 15.5 & 16.0 & 12.4 \\
         & Top-5 & 29.7 & 40.0 & 27.0 & 32.3 & 33.8 & 31.0 & 23.8 & 32.2 & 40.5 & 43.5 & 33.4 \\
\midrule
NeurBridge
      & Top-1 & 23.2 & 21.2 & 13.2 & 17.0 & 14.5 & 25.0 & 15.3 & 20.1 & 13.7 & 27.2 & 19.0 \\
      & Top-5 & 52.4 & 49.3 & 36.5 & 45.3 & 37.7 & 55.0 & 45.1 & 44.9 & 36.5 & 56.3 & 45.9 \\
      
\midrule
\bf{HCF(Ours)}
      & Top-1 & \bf 28.5 & \bf 22.5 & \bf 20.0 & \bf 20.5 & \bf 18.5 & \bf 29.5 & \bf 20.0 & \bf 20.0 & \bf 21.0 & \bf 34.0 & \bf 23.4 \\
      & Top-5 & \bf 61.0 & \bf 56.5 & \bf 45.5 & \bf 54.0 & \bf 46.5 & \bf 60.5 & \bf 53.5 & \bf 53.5 & \bf 48.5 & \bf 69.0 & \bf 54.9 \\
\bottomrule
\end{tabular}
\end{table*}

\section{Results}
\subsection{Datasets and Preprocessing}
We conduct our experiments on the THINGS-EEG dataset\cite{grootswagers2022human}, a large-scale EEG benchmark built upon the Rapid Serial Visual Presentation (RSVP) paradigm. The dataset contains recordings from 10 subjects who viewed natural object images under rapid sequential presentation.

The training set comprises 1,654 object concepts, each associated with 10 distinct images. For each subject, every image is repeated four times, yielding rich within-subject repetitions for signal averaging. The test set consists of 200 object concepts, each represented by a single image, with 80 repetitions per image per subject.

We follow the preprocessing protocol adopted in recent studies\cite{li2024visual,Wu2025BridgingTV,Zhang2025NeuroBridgeBS} to ensure fair comparison. Specifically, EEG signals are segmented into 0-1,000 ms trials following stimulus onset and baseline-corrected using the $-200$-0 ms pre-stimulus interval. The data are then downsampled to 250 Hz and normalized using multivariate noise normalization (MVNN). Finally, repetitions are averaged to enhance the signal-to-noise ratio (SNR),yielding stable sample representations.

\subsection{Implementation Details}
All experiments are implemented in PyTorch and conducted on a single RTX 4090 GPU. We use the AdamW optimizer with a weight decay of 1e-4 and a learning rate of 1e-4.

For the EEG encoder, we adopt EEGProject\cite{Wu2025BridgingTV} as the neural encoder in the intra-subject setting, and TSConv\cite{song2023decoding} in the inter-subject setting. 

For the visual encoder, we adopt the visual branch of multiple CLIP models, including RN50, RN101, ViT-B/16, ViT-B/32, ViT-L/14, and ViT-g/14. All pretrained weights are initialized from OpenCLIP\cite{cherti2023reproducible}. 

\subsection{Comparison with Baselines}
We conduct image–EEG retrieval experiments on the THINGS-EEG dataset and compare our method with several representative neural decoding approaches, including BraVL\cite{du2023decoding}, NICE\cite{song2023decoding}, ATM\cite{li2024visual}, CognitionCapturer\cite{zhang2025cognitioncapturer}, Neural-MCRL\cite{li2025neural}, VE-SDN\cite{chen2024visual}, UBP\cite{Wu2025BridgingTV}, and NeuroBridge\cite{Zhang2025NeuroBridgeBS}.

Under the intra-subject setting, the model is trained and evaluated on data from the same subject. As shown in \cref{tab:main_results}, compared with NeuroBridge, which currently achieves the best performance among existing methods, HCF improves the average Top-1 and Top-5 accuracy by 21.4\% and 8.3\%, respectively. Moreover, the improvement is consistent across all subjects, indicating stable gains in subject-specific scenarios.

Under the inter-subject setting, we adopt a leave-one-subject-out protocol, where the model is tested on an unseen subject. Due to substantial inter-subject variability in EEG signals, this setting is considerably more challenging. Nevertheless, HCF achieves the best performance on both Top-1 and Top-5 metrics, further demonstrating its generalization ability across different individuals.

\subsection{EEG-Visible Layer Analysis}
\begin{figure}[h]
  \centering
  \includegraphics[width=0.8\textwidth]{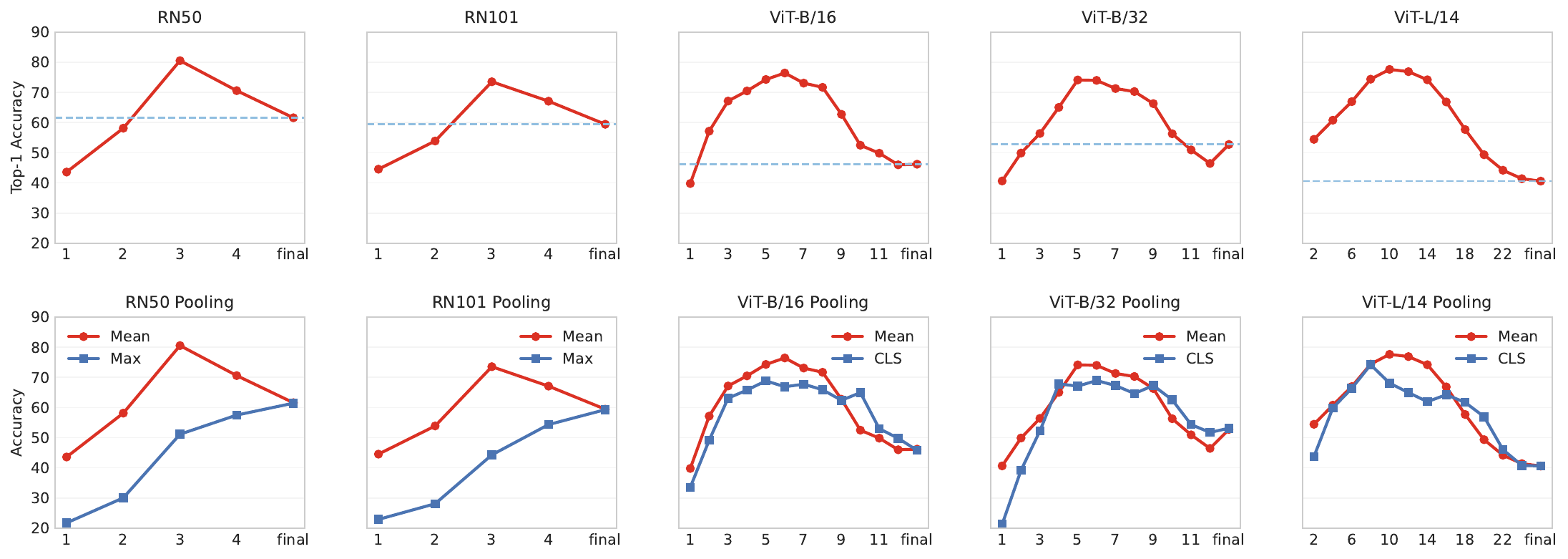}
  \caption{Layer-wise Performance Across Architectures and Pooling Strategies}
  \label{fig:layeranalysis}
\end{figure}
Based on the discussion above, the visibility of visual information in EEG signals is not uniform. Structural information associated with low spatial frequencies is generally more stable in EEG, whereas high spatial frequency details and high-level semantic information are less reliably reflected. To examine whether this phenomenon is general, we systematically evaluate retrieval performance across different layers of multiple visual backbones. The final-layer features are directly taken from the original output of the visual encoder. For non-final layers, the feature maps are converted into vector representations using the pooling operations described in \cref{sec:layer_selection}.

As shown in \cref{fig:layeranalysis} (top), both ResNet architectures (RN50, RN101) and Vision Transformer architectures (ViT-B/16, ViT-B/32, ViT-L/14) exhibit a consistent inverted U-shaped trend across layers. The middle layers achieve the best performance, while shallow and final layers perform worse. This result supports our hypothesis. The same trend appears across different pretrained backbones, indicating that the phenomenon is architecture-agnostic and general. It also provides guidance for layer selection and multi-layer fusion.

Furthermore, \cref{fig:layeranalysis} (bottom) shows the impact of different pooling strategies. For ResNet, mean pooling consistently outperforms max pooling at all layers. For ViT, mean pooling also performs better than CLS token pooling, with the most stable advantage around the middle layers. These results suggest that, for EEG–vision alignment, more uniform aggregation of global information is better suited for extracting stable structural representations. In contrast, max pooling emphasizes local peak responses, and the CLS token tends to encode globally semantic information shaped by pretraining objectives. Both may amplify high-frequency details or overly abstract semantics that EEG signals cannot reliably capture, thereby weakening cross-modal alignment.

\subsection{Multi-layer Feature Fusion across Different Visual Backbones}

\begin{figure}[!t]
\centering

\begin{minipage}{0.4\textwidth}
    \centering
    \includegraphics[width=\linewidth]{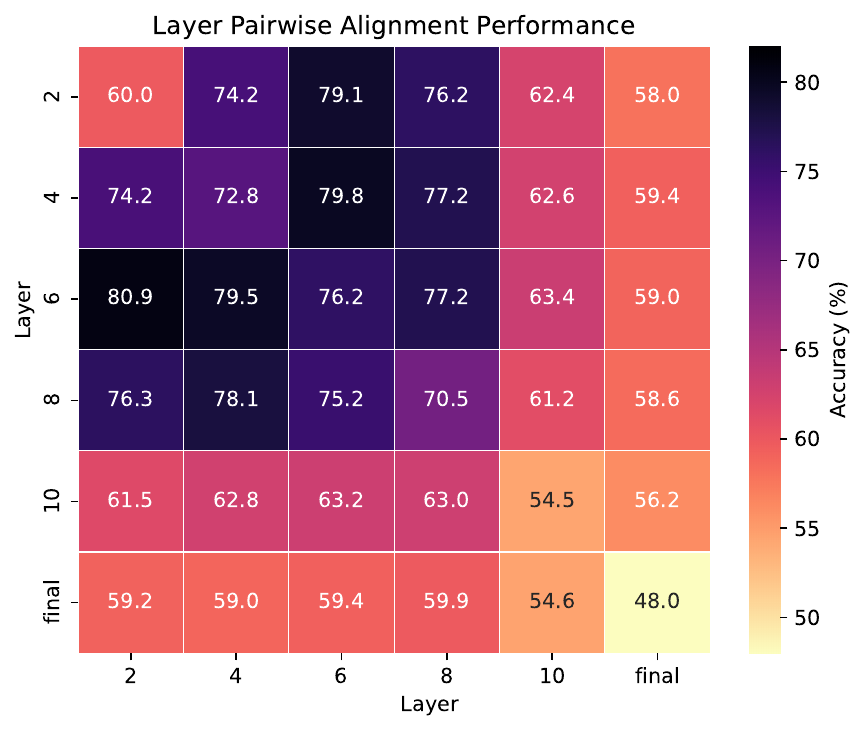}
    \caption{Layer Pairwise Fusion Analysis (ViT)}
    \label{fig:heatmap}
\end{minipage}
\hfill
\begin{minipage}{0.4\textwidth}
    \centering
    \includegraphics[width=\linewidth]{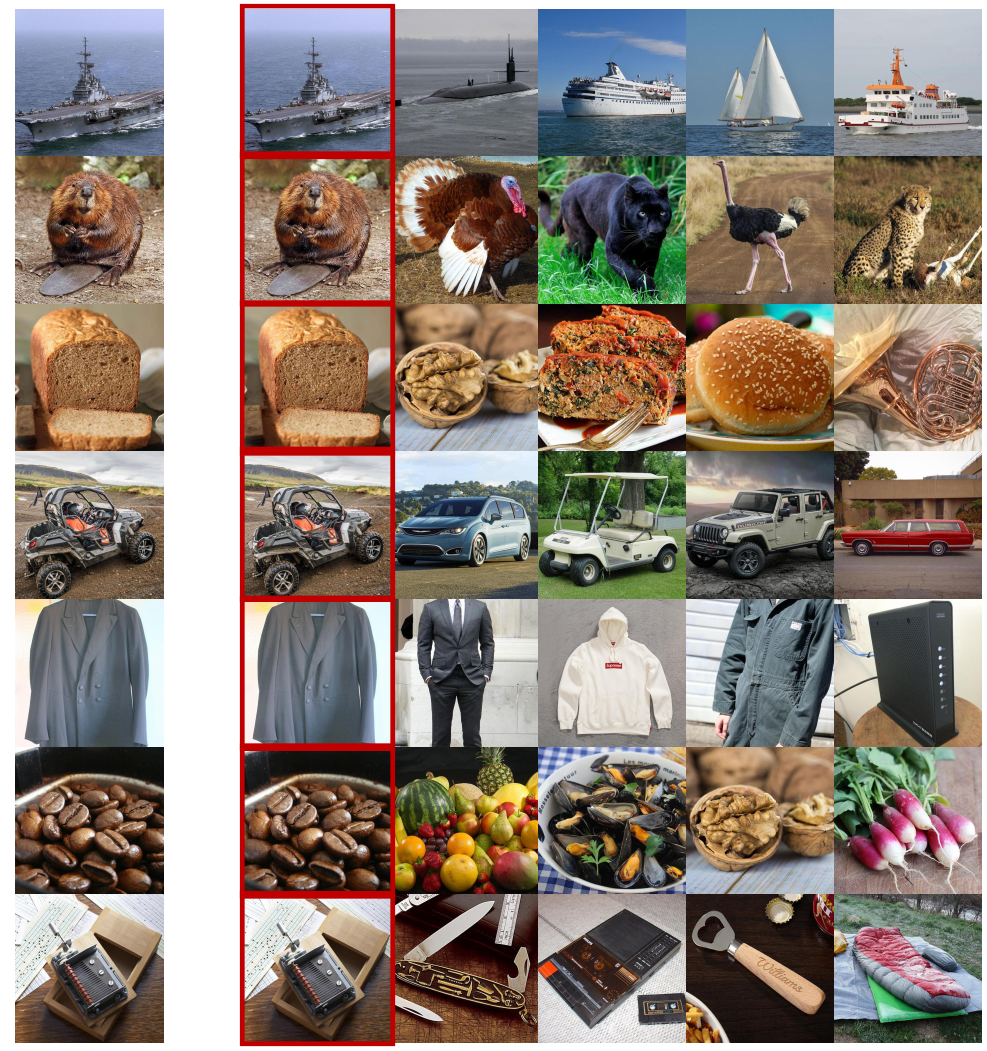}
    \caption{Retrieved Samples}
    \label{fig:retrieved}
\end{minipage}

\end{figure}

To further investigate the complementary relationships among features from different hierarchical levels, we systematically evaluate the fusion of different layer combinations across multiple visual backbones. Our experiments show that, for both ResNet and ViT architectures, fusing two carefully selected layers consistently outperforms both single-layer features and the fusion of more layers, suggesting that appropriate layer selection is more critical than simply aggregating more layers.

Specifically, for each backbone we select two visual layers and fuse their representations through feature concatenation. The fused features are then aligned with EEG embeddings using a unified contrastive learning framework, allowing us to evaluate how different layer combinations influence EEG-image retrieval performance.

Through systematic experiments, we observe that different visual backbones exhibit distinct optimal layer fusion strategies. For ResNet-based models (ResNet50 and ResNet101), the best performance is typically achieved by combining an intermediate layer with the final layer. Compared with single-layer representations, this combination significantly improves retrieval accuracy, suggesting that the structural information captured in intermediate layers and the high-level semantic information encoded in the final layer provide complementary cues.

However, a different pattern emerges for Vision Transformer models (ViT-B/16, ViT-B/32, and ViT-L/14). As shown in \cref{fig:heatmap}, the best results are usually obtained by fusing intermediate layers with each other, rather than involving the final layer. This observation suggests that in ViT models, deeper layers tend to encode more abstract semantic representations, whose visibility in EEG signals is relatively limited.



This difference can be explained by the hierarchical characteristics of CNN and Transformer architectures. In CNN-based models, the final layers still retain certain spatial structural cues inherited from earlier layers. Therefore, combining intermediate and final layers allows a balance between structural information and semantic abstraction. In contrast, in Vision Transformers, due to the global self-attention mechanism and large-scale vision–language pretraining, deeper layers tend to encode increasingly semantic and category-level representations, gradually discarding structural details. These highly abstract semantic representations tend to be less stable and less visible in EEG signals, making them less suitable for cross-modal alignment.

In comparison, intermediate layers of ViT models still preserve richer structural and shape-related information, which better matches the EEG-visible representations observed in neural signals. Consequently, fusing intermediate layers leads to more stable and effective EEG–vision alignment.

Overall, these findings suggest that the effectiveness of multi-layer feature fusion does not depend on incorporating more layers, but rather on selecting hierarchical representations that align with EEG-visible information. This observation further supports our Neural Visibility hypothesis, which posits that EEG signals more reliably encode structural visual information, while highly abstract semantic representations exhibit lower readability in EEG.

\subsection{Spatial Frequency Analysis of EEG Visibility}

\begin{figure}[!t]
  \centering
  \includegraphics[width=0.8\textwidth]{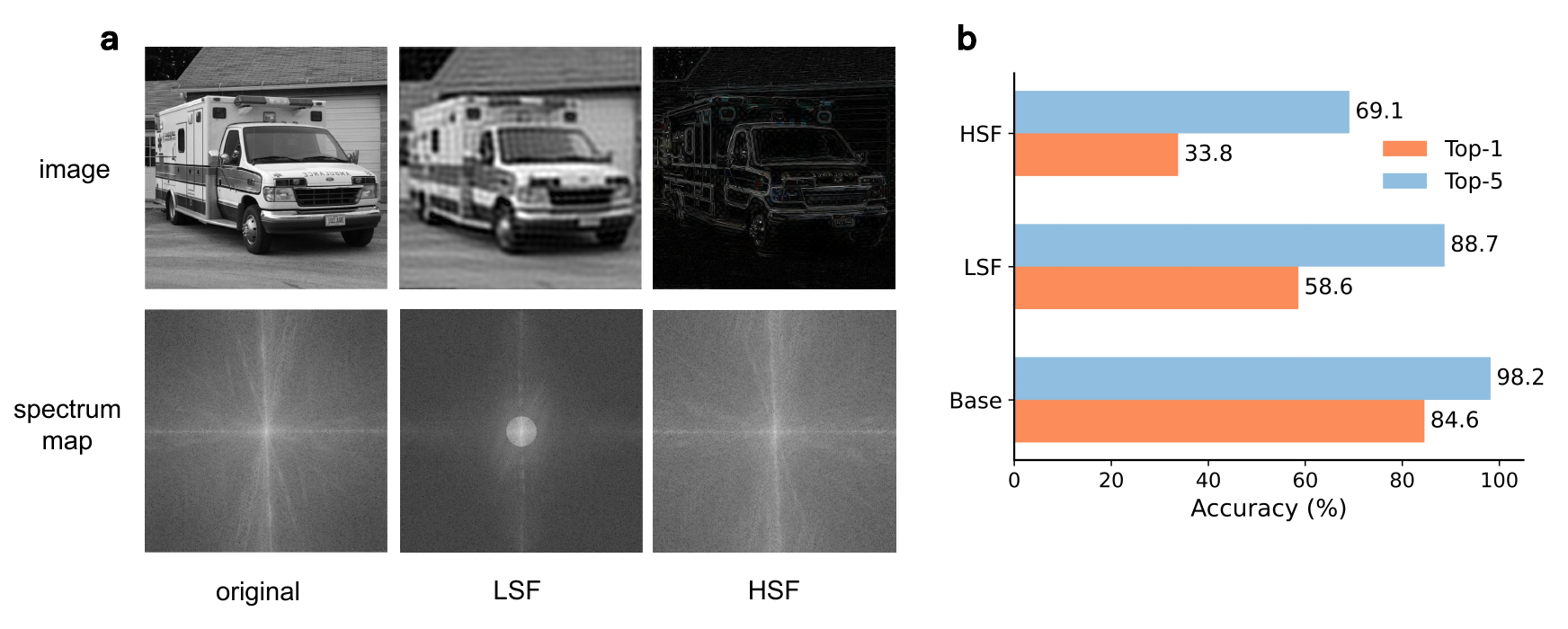}
  \caption{Visualization of spatial frequency components and their impact on EEG–image retrieval performance.}
  \label{fig:frequency}
\end{figure}

To further analyze the visibility of different visual information in EEG representations, we conduct a frequency decomposition experiment. We apply a low-pass filter (LPF) and a high-pass filter (HPF) with a cutoff ratio of 0.2 to decompose the original images into low spatial frequency (LSF) components and high spatial frequency (HSF) components, as shown in \cref{fig:frequency}(a). The models are then trained and evaluated within the HCF framework using these filtered images.

The results are shown in \cref{fig:frequency}(b). When images are processed with the high-pass filter, which mainly preserves edge and texture details, the model performance drops sharply, with the Top-1 accuracy decreasing by 50.8\% and the Top-5 accuracy decreasing by 29.1\%. In contrast, the LPF model still maintains strong alignment ability and clearly outperforms the HPF setting. These results suggest that EEG representations align better with low spatial frequency structural information than with high-frequency details.

\subsection{Generalization across EEG Encoders}
\begin{table}[!t]
\centering
\caption{Generalization across EEG encoders on THINGS-EEG}
\label{tab:encoder_generalization}

\scriptsize
\setlength{\tabcolsep}{4pt}

\begin{tabular}{l cccccc cc}
\toprule

& \multicolumn{2}{c}{Final}
& \multicolumn{2}{c}{Low}
& \multicolumn{2}{c}{HCF(Ours)}
& \multicolumn{2}{c}{Gain(\%)} \\

Encoder
& top-1 & top-5
& top-1 & top-5
& top-1 & top-5
& top-1 & top-5 \\

\midrule

ATM
& 40.15 & 74.45
& 75.20 & 95.85
& 79.91 & 96.77
& \bf +99.0 & \bf +29.98 \\

EEGConformer
& 6.60 & 23.15
& 5.90 & 22.50
& 7.65 & 27.90
& \bf +15.9 & \bf +20.52 \\

EEGNetV4
& 30.90 & 62.50
& 67.85 & 93.00
& 71.00 & 93.20
& \bf +129.8 & \bf +49.12 \\

ShallowFBCSP
& 18.50 & 44.40
& 40.70 & 73.85
& 41.80 & 75.90
& \bf +125.9 & \bf +70.95 \\

\bottomrule
\end{tabular}

\end{table}

To evaluate the generalization ability of our method, we apply it to several representative EEG encoders, including ATM \cite{li2024visual}, EEGConformer \cite{song2022eeg}, EEGNetV4 \cite{lawhern2018eegnet}, and ShallowFBCSP \cite{schirrmeister2017deep}. 

In all experiments, RN50 is used as the visual backbone. We adopt the best-performing feature settings identified in previous experiments: Low uses features from layer3, while Concat fuses features from layer3 and the final layer. The results are shown in \cref{tab:encoder_generalization}. Our method consistently improves performance across different EEG encoders. For example, on ATM, compared with using the final-layer features, Concat achieves relative improvements of +99.0\% in Top-1 accuracy and +29.98\% in Top-5 accuracy. Similar trends are observed for EEGNetV4 and ShallowFBCSP, where Top-1 accuracy improves by +129.8\% and +125.9\%, respectively.

These results indicate that the proposed method does not rely on a specific network architecture and can serve as a general alignment strategy with strong generalization ability. The findings further support our core claim: compared with final-layer representations, EEG signals exhibit higher visibility to mid-level structural features, and multi-layer feature fusion can better match the visual information encoded in EEG signals.

\subsection{Ablation on Visual Backbone and Layer Composition}

\begin{table}[!t]
\centering
\caption{Ablation on visual backbones and feature composition on THINGS-EEG.}
\label{tab:ablation}

\scriptsize
\setlength{\tabcolsep}{3pt}

\begin{tabular}{lcccccccccc}
\toprule

& \multicolumn{2}{c}{ResNet50}
& \multicolumn{2}{c}{ResNet101}
& \multicolumn{2}{c}{ViT-B/16}
& \multicolumn{2}{c}{ViT-B/32}
& \multicolumn{2}{c}{ViT-L/14} \\

\cmidrule(lr){2-3}
\cmidrule(lr){4-5}
\cmidrule(lr){6-7}
\cmidrule(lr){8-9}
\cmidrule(lr){10-11}

& top-1 & top-5
& top-1 & top-5
& top-1 & top-5
& top-1 & top-5
& top-1 & top-5 \\

\midrule

final
& 61.6 & 89.8
& 59.5 & 87.9
& 46.2 & 78.8
& 52.8 & 83.9
& 40.6 & 73.4 \\

$\Delta$
& {\scriptsize\textbf{+19.3}} & {\scriptsize\textbf{+7.1}}
& {\scriptsize\textbf{+14.1}} & {\scriptsize\textbf{+6.8}}
& {\scriptsize\textbf{+30.3}} & {\scriptsize\textbf{+17.3}}
& {\scriptsize\textbf{+21.4}} & {\scriptsize\textbf{+11.4}}
& {\scriptsize\textbf{+37.1}} & {\scriptsize\textbf{+23.0}} \\

mid
& 80.9 & 96.9
& 73.6 & 94.7
& 76.5 & 96.1
& 74.2 & 95.3
& 77.7 & 96.4 \\

$\Delta$
& {\scriptsize\textbf{+3.7}} & {\scriptsize\textbf{+1.3}}
& {\scriptsize\textbf{+2.3}} & {\scriptsize\textbf{+1.1}}
& {\scriptsize\textbf{+3.4}} & {\scriptsize\textbf{+0.8}}
& {\scriptsize\textbf{+5.0}} & {\scriptsize\textbf{+1.5}}
& {\scriptsize\textbf{+3.3}} & {\scriptsize\textbf{+0.9}} \\

HCF
& 84.6 & 98.2
& 75.9 & 95.8
& 79.9 & 96.9
& 79.2 & 96.8
& 81.0 & 97.3 \\

\bottomrule
\end{tabular}

\end{table}

We conduct an ablation study by varying the visual backbone (ResNet / ViT) and the layer composition strategy (Final / Low / HCF). As shown in \cref{tab:ablation}, replacing the final-layer features with intermediate-layer features consistently improves retrieval performance, and fusing multi-layer features brings further gains. These results indicate that the improvement of our method is not tied to a specific backbone, but mainly comes from the proposed EEG-Visible Layer Selection Strategy and Hierarchically Complementary Fusion framework.

\section{Conclusion}

In this paper, we introduce the concept of Neural Visibility into EEG-based brain decoding and propose the EEG-Visible Layer Selection Strategy. This strategy aligns EEG signals with intermediate-layer visual features to minimize the cross-modal information mismatch between neural activity and visual stimuli. Furthermore, we propose a novel Hierarchically Complementary Fusion (HCF) framework that jointly and dynamically adjusts multi-level visual representations to achieve more precise brain-vision feature alignment. Experiments on the THINGS-EEG dataset demonstrate 84.6\% zero-shot accuracy (+21.4\% improvement) and up to 129.8\% relative gains across diverse baseline encoders. These results effectively expand the application boundaries of universal EEG-based BCIs.




%
%
\bibliographystyle{splncs04}
\bibliography{main}
\end{document}